\documentclass[10pt,twocolumn,letterpaper]{article}

\usepackage[pagenumbers]{cvpr} 

\usepackage{times}
\usepackage{epsfig}
\usepackage{graphicx}
\usepackage{amsmath}
\usepackage{amssymb}

\usepackage[ruled,linesnumbered]{algorithm2e}
\usepackage{algorithm2e}
\usepackage{booktabs}
\usepackage{float}  
\usepackage{multirow}
\usepackage{caption}
\usepackage{soul}
\usepackage{dsfont}
\usepackage{enumitem}

\usepackage{color}
\usepackage{amsthm}

\usepackage[pagebackref,breaklinks,colorlinks]{hyperref}

\usepackage[capitalize]{cleveref}
\crefname{section}{Sec.}{Secs.}
\Crefname{section}{Section}{Sections}
\Crefname{table}{Table}{Tables}
\crefname{table}{Tab.}{Tabs.}

\def\cvprPaperID{257} 
\def\confName{CVPR}
\def\confYear{2023}

\makeatletter
\newbox\abstract@box
\renewenvironment{abstract}
  {\global\setbox\abstract@box=\vbox\bgroup
     \hsize=\textwidth\linewidth=\textwidth
    \small
    \begin{center}%
    {\bfseries \abstractname\vspace{-.5em}\vspace{\z@}}%
    \end{center}%
    \quotation}
  {\endquotation\egroup}
\expandafter\def\expandafter\@maketitle\expandafter{\@maketitle
  \ifvoid\abstract@box\else\unvbox\abstract@box\if@twocolumn\vskip1.5em\fi\fi}
\makeatother

\begin{document}
\title{Unimodal Training-Multimodal Prediction: Cross-modal Federated Learning with Hierarchical Aggregation}
\author{Rongyu Zhang\textsuperscript{\rm 1,4}\thanks{Equal contribution: rongyuzhang@link.cuhk.edu.cn}, 
Xiaowei Chi\textsuperscript{\rm 2*},
Guiliang Liu\textsuperscript{\rm 1},
Wenyi Zhang\textsuperscript{\rm 3},\\
Yuan Du\textsuperscript{\rm 4},
Fangxin Wang\textsuperscript{\rm 1}\thanks{Corresponding author: wangfangxin@cuhk.edu.cn}\\
\textsuperscript{\rm 1}The Chinese University of Hong Kong, Shenzhen, \textsuperscript{\rm 2}The Chinese University of Hong Kong, \\
\textsuperscript{\rm 3}University of California, Irvine,
\textsuperscript{\rm 4}Nanjing University\\ 
}


\begin{abstract}
Multimodal learning has seen great success mining data features from multiple modalities with remarkable model performance improvement. Meanwhile, federated learning (FL) addresses the data sharing problem, enabling privacy-preserved collaborative training to provide sufficient precious data. Great potential, therefore, arises with the confluence of them, known as multimodal federated learning. However, limitation lies in the predominant approaches as they often assume that each local dataset records samples from all modalities. In this paper, we aim to bridge this gap by proposing an \texttt{Unimodal Training - Multimodal Prediction} (UTMP) framework under the context of multimodal federated learning. 
We design \texttt{HA-Fedformer}, a novel transformer-based model that empowers unimodal training with only a unimodal dataset at the client and multimodal testing by aggregating multiple clients' knowledge for better accuracy. The key advantages are twofold. Firstly, to alleviate the impact of data non-IID, we develop an uncertainty-aware aggregation method for the local encoders with layer-wise Markov Chain Monte Carlo sampling. Secondly, to overcome the challenge of unaligned language sequence, we implement a cross-modal decoder aggregation to capture the hidden signal correlation between decoders trained by data from different modalities. Our experiments on popular sentiment analysis benchmarks, CMU-MOSI and CMU-MOSEI, demonstrate that \texttt{HA-Fedformer} significantly outperforms state-of-the-art multimodal models under the UTMP federated learning frameworks, with 15$\%$-20$\%$ improvement on most attributes.
\end{abstract}
\maketitle
\section{Introduction}
Human perception of the world usually consists of information with multiple modalities, including sounds, texts, images, etc. 
With the continuous development of artificial intelligence, multimodal learning has gradually become the focus of attention. However, although multimodal learning can improve inference accuracy, many users still fail to get satisfactory prediction accuracy due to the limited training data. To surmount the problem of poor model performance caused by insufficient user data, a paradigm of federated learning (FL)~\cite{mcmahan2017communication} with multi-client co-training a global model in a privacy-preserving manner has been applied to multimodal learning.

\begin{figure}[t]
\centering
\includegraphics[scale=0.23]{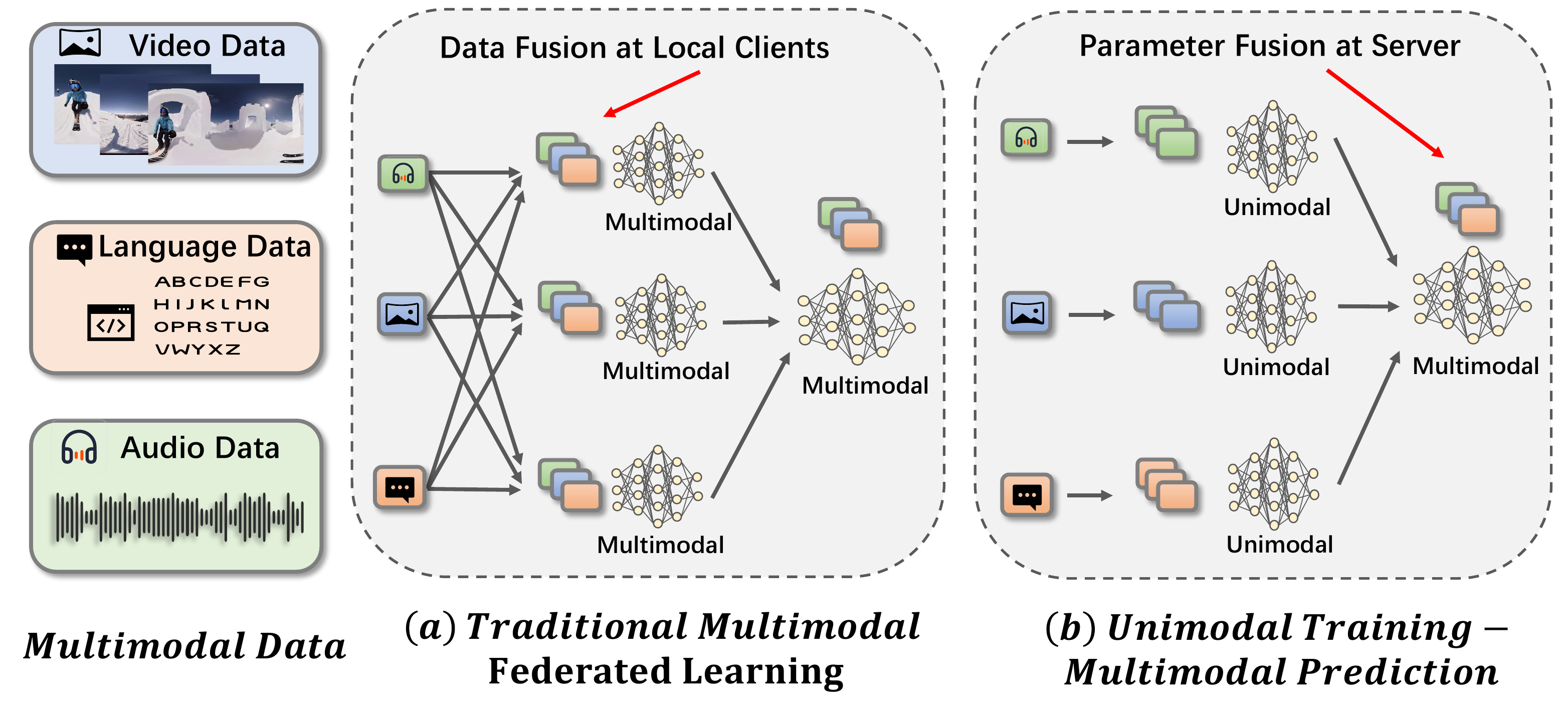}
\caption{Comparison of (a) Traditional multimodal federated learning taking multimodal data and (b) UTMP taking unimodal data as training input.} 
\vspace{-1em}
\label{clustering}
\end{figure}

Existing multimodal federated learning methods~\cite{liang2020think,chen2022towards} are based on an implicit assumption over the congruity of data modalities across different clients,
however, in practice, many clients only have sensors that can collect unimodal data. 
For example, many IoT devices (e.g., smart speakers) only collect audio data through conversations, and surveillance cameras collect video data due to noisy environments. Since it is generally difficult to determine data ownership (as sensors are indiscriminate in collecting data and data transfer may lead to data leakage), it is infeasible to collect and align the data of different modalities for supporting central training. It requires an FL solution to answer the following question: \textbf{Can a multimodal model be trained via clients with access to only unimodal data as showed in Figure \ref{clustering}?}
Although Chen et al.~\cite{chen2022fedmsplit} proposed FedMSplit that studies the occasional missing modality while training. 
To answer this question, we propose a UTMP framework that puts a step forward: training a multimodal model with only unimodal data in FL.


In this paper, we formally define the aforementioned scenario as a \texttt{Unimodal Training -Multimodal Prediction} (UTMP) framework. Such a framework allows users with only \textit{unimodal} data to participate in the \textit{multimodal} federated learning, providing a more comprehensive application scope for multimodal learning.
However, learning a global multimodal model under the UTMP setting raises new challenges that require a novel solution for several reasons:
\textbf{Data non-IID:} our empirical study shows a difficulty that the local models are inclined to {\it overfit the unimodal data distribution} under the UTMP setting, especially for the data with simple representations (e.g., text). 
\textbf{Data Unalignment:} an essential prerequisite for multimodal learning is creating the data alignment across modalities. 
However, directly aligning data under the UTMP setting is impossible since the data cannot be communicated across clients. 

We foray into uncharted territory to tackle the above-mentioned challenges by proposing Hierarchical Aggregated Multimodal Federated Transformer (\textit{HA-Fedformer}) with a tailored model aggregation strategy.
Specifically, \textit{HA-Fedformer} consists of a transformer-based encoder for each modality and one shared decoder for cross-modality feature fusion. 
Such structure provides a prerequisite for our UTMP framework. By concatenating different unimodal-trained encoders during model aggregation, we can obtain a multimodal model with full-modal encoders, capable of making predictions with multimodal data for higher accuracy. 
However, due to the two challenges mentioned above, aggregating encoders and decoders directly through traditional FL methods cannot achieve satisfactory results. To this end, we propose a hierarchical aggregation method for encoders and decoders, respectively. 
\textbf{Solution 1: PbEA (Posterior-based Encoder Aggregation)} Uncertainty is a metric measuring if the network \textit{knows what it don't knows}~\cite{gal2016dropout}. We consider the layer-wise uncertainty of each local model and aggregate encoders based on the mean and variance of their posterior inferred with Markov Chain Monte Carlo sampling to alleviate data non-IID problem.
\textbf{Solution 2: CmDA (Cross-modal Decoder Aggregation)} We also implement a cross-modal aggregation method to find signal correlations between different modalities implicit in the model weights of decoders trained on data from different modalities to achieve alignment in model parameter level instead of feature level.


We evaluate \textit{HA-Fedformer} against two widely-used multimodal sentiment analyses datasets: CMU-MOSI and CMU-MOSEI, to prove that our proposed \textit{HA-Fedformer} can overcome the difficulties of data non-IID and unalignment with only single unimodal data input during local training and inference with sufficient multimodal data for higher inference accuracy. 
The main contributions of this paper can be summarized as the following:
\begin{itemize}
\setlength\itemsep{0em}
    \item[-] First, we define \textit{Unimodal Training - Multimodal Prediction} framework in the context of multimodal federated learning, which greatly expands its application scope.

    \item[-] Second, we proposed a hierarchical model aggregation method with Posterior-based Encoder Aggregation and Cross-modal Decoder Aggregation for \textit{HA-Fedformer}, overcoming the data non-IID and sequence unalignment challenges in UTMP.

    \item[-] Third, to our best knowledge, our work Hierarchical Aggregated Multimodal Federated Transformer (\textit{HA-Fedformer}) is the first to achieve UTMP with only unimodal data locally trained. 

\end{itemize}

\section{Related works}
\subsection{Federated learning}
 Federated learning is proposed ~\cite{mcmahan2017communication} to protect user privacy as a critical learning scenario in large-scale applications. 
 Many works ~\cite{karimireddy2020scaffold, woodworth2020local, woodworth2020minibatch,li2019convergence,li2021federated} have demonstrated that none-independent and identically distributed (non-IID) data brought by heterogeneous users have a significant adverse effect on the convergence and accuracy of the traditional aggregation strategies~\cite{reddi2020adaptive, khodak2019adaptive}. 
 Ji et al. proposed FedAtt~\cite{wang2020optimizing}, which aggregates model updates from the updates with biased weights in order to train models with higher qualities. 
 Liang et al. ~\cite{liang2020think} jointly optimized mixed global and local models to seek a trade-off between overfitting and generalization.
 Boughorbel et al. ~\cite{boughorbel2019federated} also introduced an uncertainty-aware learning algorithm into federated learning for model aggregation, providing a new perspective for the aggregation of network parameters. 
 

\subsection{Multimodal learning}

Multi-modality in learning analytics and learning science is under the spotlight, especially for human language processing. 
Since human language contains time series, analyzing human language requires synthesizing and fusing time-varying signals. ~\cite{liang2018multimodal, tsai2018learning}. 
Many advanced models ~\cite{yu2021learning, sun2020learning, wang2019words, pham2019found} have been proposed but are very dependent on the context information in the short term and can only capture the relationship between the various modes on the aligned multimodal data. 
With the introduction of Transformer ~\cite{vaswani2017attention}, many researchers ~\cite{zolfaghari2021crossclr, li2020unimo, nagrani2021attention}have proposed a cross-attention fusion mechanism between modal vectors by borrowing its self-attention mechanism and has proved that cross-attention mechanism has an extended scale for unaligned language sequences. 


\subsection{Multimodal federated learning}

Currently, most federated learning frameworks are based on single-modal data classification or recognition tasks, and only a few work on multimodal federated learning tasks. Chen et al. ~\cite{chen2022towards} proposed hierarchical gradient blending (HGB) to alleviate these inconsistencies in collaborative learning. Zhao et al. ~\cite{zhao2022multimodal} proposed a multimodal and semi-supervised federated learning framework that trains auto-encoders to extract shared or correlated representations from different local data modalities on clients. 
Chen et al.~\cite{chen2022fedmsplit} proposed FedMSplit considering a scenario similar to UTMP by constructing a dynamic and multi-view graph structure to adaptively select multimodal client models where some modality may be missing. Yang et al.~\cite{yang2022cross} also considered taking unimodal data in specific Human Activity Recognition FL tasks. However, their methods take unimodal data for testing, which potentially undermines the model's performance.


\section{Problem definition}

\begin{figure*}[t]
\centering
\includegraphics[scale=0.2]{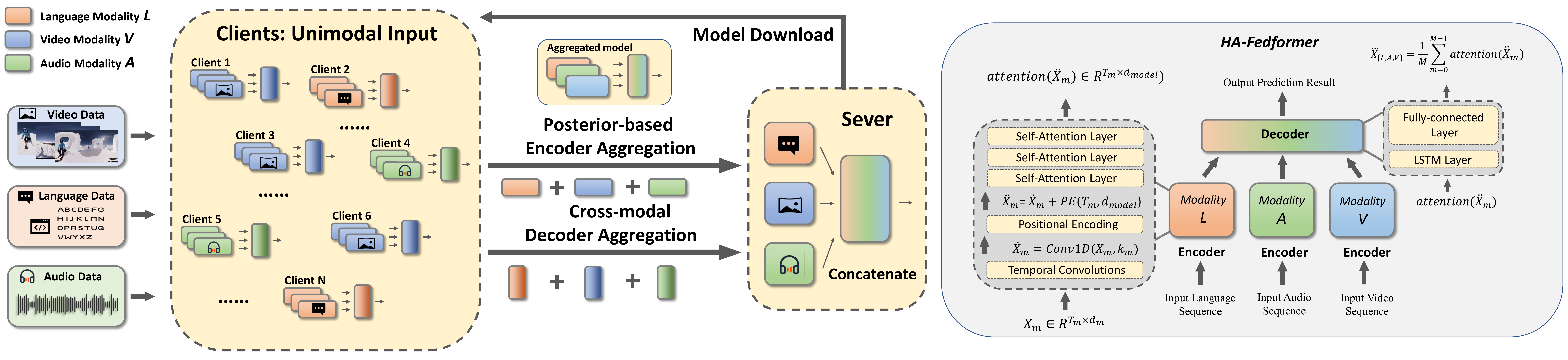}
\caption{UTMP framework and the construction of HA-Fedformer. Each transformer-based encoder extracts data features from one modality (L, A, or V), and their training data are unaligned with the non-IID distribution. These clients' models will be aggregated to a multimodal model on the server by PbEA and CmDA. Then, the merged global model will be sent back to each client for local training.}
\vspace{-1em}
\label{framework}
\end{figure*}

\subsection{Preliminaries}

Federated learning aims at training a global model by utilizing the dataset stored at local clients. Each client stores a local dataset $\mathcal{D}_{k}$ where 1) $k\in\{1,\dots,K\}$ denotes the $k^{th}$ client and 2) the inputs $\boldsymbol{X}$ and their labels $\boldsymbol{y}$ are sampled from a local data distribution $(\mathcal{X},\mathcal{Y})^{k}$. At communication round $t$, the local model parameter $\boldsymbol{\theta}_{k}^{t}$ can be updated by stochastic gradient descent(SGD):

\begin{equation}
\begin{aligned}
\boldsymbol{\theta}_{k}^{t+1} &= \boldsymbol{\theta}^{t}_{k} - \varphi\nabla\ell_{k}(\upsilon(\boldsymbol{X};\boldsymbol{\theta}^{t}_{k}),\boldsymbol{y}) 
\end{aligned}
\end{equation}
where $\upsilon$ denotes a model parameterized by $\boldsymbol{\theta}_{k}^{t}$, $\varphi$ is the learning rate, and $\ell(\cdot)$ is a user-specific loss function (e.g., Mean Square Error loss). Thus, the final optimization problem of FL can be formulated as follows:

\begin{equation}
\min\limits_{\boldsymbol{\theta}}\Big\{F(\boldsymbol{\theta}) = \sum_{k=1}^{N}\alpha_{k}f_{k}(\boldsymbol{\theta})\Big\}
\end{equation}
where $\alpha_{k}$ is the weight for client $k$. We assume the objective function $f_{k}$ is \textit{convex} and \textit{L-smooth}~\cite{chen2022towards}.

\subsubsection{Vanilla multimodal federated
learning} 


Traditional multimodal FL algorithms assume a client has access to the data from all modalities.
Let $\rho_{k}(\boldsymbol{X},\boldsymbol{y})$ defines the density of a data point in dataset $\mathcal{D}_{k}$ ($k\in[1,K]$ denotes client number.). Since $\mathcal{D}_{k}$ stores data from all modality, $\rho_{k}(\boldsymbol{X},\boldsymbol{y})=\sum_{m=1}^{M}p_{k}(m)\rho_{m}(\boldsymbol{X},\boldsymbol{y})$ where $\rho_{m}(\boldsymbol{X},\boldsymbol{y})$ denotes the density of dataset for the $m^{th}$ modality and $p_{k}(m)\in[0,1]$ is a mixing coefficient that merges unimodal densities to a client density.
Under this setting, these algorithms can learn a {\it local multimodal models} at each client with the objective:

\begin{equation}
    f_{k}=\frac{1}{|D_{k}|}\sum_{(\boldsymbol{X},\boldsymbol{y})\in D_{k}}\oplus_{m=1}^{M}\mathds{1}_{(\boldsymbol{X},\boldsymbol{y})\in\mathcal{D}_{m}}\ell\left[\upsilon_{m,k}(\boldsymbol{X};\boldsymbol{\theta}_{k}),\boldsymbol{y}\right]
\end{equation}
where $\upsilon_{m,k}$ denotes the model for the modality $m$ at client $k$, $\oplus$ denotes loss aggregation across modality $1$ to $M$ and $\mathds{1}_{(\boldsymbol{X},\boldsymbol{y})\in\mathcal{D}_{m}}$ identifies whether the data $(\boldsymbol{X},\boldsymbol{y})$ is sampled from $\mathcal{D}_{m}$ (as a part of our prior knowledge, this identifier is known before training).
Given the pre-trained local model parameters $\{\boldsymbol{\theta}_{k}\}_{k=1}^{K}$, the global model aggregates them into a global one for better prediction accuracy. Note that this process {does not involve cross-modality aggregation} since the local models are trained with multi-modal data.







\subsection{Unimodal Training - Multimodal Prediction}
In this work, we focus on a more challenging \textit{Unimodal Training - Multimodal Prediction} (UTMP) framework.
UTMP enables clients with only unimodal data to participate in federated learning. The algorithms must build a global model by utilizing knowledge learned by local clients. Ideally, the global model can predict multimodal data with better performance than local models. 

 Under the UTMP framework, each client stores only data from a single modality, so $\rho_{k}(\boldsymbol{X},\boldsymbol{y})=\alpha_{\xi}\rho_{m}(\boldsymbol{X},\boldsymbol{y})$ where $\alpha_{\xi}$ denotes the potential distribution shift.
Therefore, the model $\upsilon_{m,k}$ at the local client $k$ takes only the data from a single modality $m$ as input. The optimization problem of \textit{local unimodal training} can be formulated as:


\begin{equation}
    f_{k}=\frac{1}{|D_{k}|}\sum_{(\boldsymbol{X},\boldsymbol{y})\in D_{k}}\ell\left[\upsilon_{m,k}(\boldsymbol{X};\boldsymbol{\theta}_{k}),\boldsymbol{y}\right]
\end{equation}

Note that UTMP requires unimodal training at the local client but cross-modal aggregation at the global server, so the global model parameters $\boldsymbol{\theta}=\oplus_{m=1}^{M}\odot_{k=1}^{K}\boldsymbol{\theta}_{k,m}$, where $\odot_{k=1}^{K}$ and $\oplus_{m=1}^{M}$ denotes single-modality and cross-modality aggregation in federated learning. This hierarchical aggregation serves as the main motivation of the proposed approach, including 1) Posterior-based Encoder Aggregation(PbEA) (corresponding to $\odot$) and 2) Cross-modal Decoder Aggregation(CmDA) (corresponding to $\oplus$).

However, moving cross-modal aggregation from clients to servers creates a more challenging problem since 1)
UTMP requires aggregating model without access to local data. However, the model parameters trained for different modalities are highly independent and thus become hard to aggregate. 2) Unlike traditional cross-modal prediction approaches, the unimodal data scattered among different clients cannot be aligned through pre-processing in federated learning. Accordingly, cross-modal information fusion cannot be achieved in local training. These difficulties make it impossible for traditional cross-modal and federated learning methods to perform model aggregation.

Despite these challenges, we believe this UTMP framework has a more close connection to real-world applications: UTMP enables edge computing devices to participate in federated learning by utilizing only the unimodal data collected by themselves. UTMP provides a broader range of applications for multimodal federated learning and thus has a more considerable impact than the previous designs.



\section{Proposed method}
In this work, we split the feature extraction and prediction layers to cope with the hierarchical aggregation. Specifically, we introduce a transformer-based encoder for feature extraction and an RNN-based decoder for prediction. Based on these structures, we propose {\it Posterior-based Encoder Aggregation} (PbEA) and {\it Cross-modal Decoder Aggregation} (CmDA) that handle the encoder and decoder parameters respectively.
The overall architecture of \textit{HA-Fedformer} can be illustrated in Figure \ref{framework}. 


\subsection{Model architecture}
\subsubsection{Encoder}
We consider three major modalities that are commonly studied in multimodal learning, including language (\textit{L}), audio (\textit{A}), and video (\textit{V}) modalities. For modality $m \in\{L,A,V \}$, the input matrix $\boldsymbol{X}_m \in \{\mathbb{R}^{T_m \times d_m}\}$ is used to denote the three input feature sequences, where $T(\cdot)$ denotes the length of the sequence and $d(\cdot)$ denotes the feature dimension of the sequence. The main components are:

{\it Temporal Convolutions.}
Although the training data of the local model belongs to different modalities, multimodal data is still required as input when making predictions. Therefore, it is necessary to reshape the input sequence to a uniform shape. We choose 1$\times$1 temporal convolution to perform the adjustment: 

\begin{equation}
\begin{aligned}
        \dot{\boldsymbol{X}}_{m} &=
    Conv1D(\boldsymbol{X}_m, \iota_{m})
\end{aligned}
\end{equation}
where $\iota_m$ denotes the convolutional kernel for modality $m$, $d_{model}$ is a customize dimension, and the output of encoders $\dot{\boldsymbol{X}}_{m} \in \mathbb{R}^{T_{m}\times d_{model}}$ are matrices with uniformed shape.

{\it Positional Encoding.}
We add positional embedding (PE) to $\dot{\boldsymbol{X}}_{m}$ for incorporating temporal information into the adjusted sequences. Following \textit{Transformer}~\cite{vaswani2017attention}, we augment PE to $\dot{\boldsymbol{X}}_{m}$: 




\begin{equation}
\begin{aligned}
        \ddot{\boldsymbol{X}}_{m} = \dot{\boldsymbol{X}}_{m} + PE(T_{m},d_{model})
\end{aligned}
\end{equation}
where $\ddot{\boldsymbol{X}}_{m}$ represents the low-level position-aware embeddings and $PE(T_{m},d_{model})$ computes the embeddings for each position index of the data sequences. We leave the details of the computation of PE in Appendix A.1.

{\it Self-Attention Transformer.}
We perform self-attention based embedding feature extraction~\cite{vaswani2017attention} for the input temporal sequences $\ddot{\boldsymbol{X}}_{m}$. As local clients only hold unimodal data, the output of the three encoders ${attention}(\ddot{\boldsymbol{X}}_{m})$ will be the embeddings of one modality. 

\subsubsection{Decoder}
The decoder is constructed with LSTM and fully-connected layers with the residual module, which is a classifier module for decoding predicted labels instead of decoding input features like auto-encoding models. 
Note that the local client has only the unimodal data, but each client maintains $M$ (the total number of modalities, e.g., three for L, A, and V) encoders for extracting features from multimodal data. During implementation, we feed $\boldsymbol{X}_{m}$ to all the encoders and use the average embedding from all encoders as the input of our decoder $\dddot{\boldsymbol{X}}_{m} =\frac{1}{M} \mathop{\sum}\limits_{m=0}^{M-1} attention(\ddot{\boldsymbol{X}}_{m})$.

\begin{figure}[t]
\centering
\includegraphics[scale=0.33]{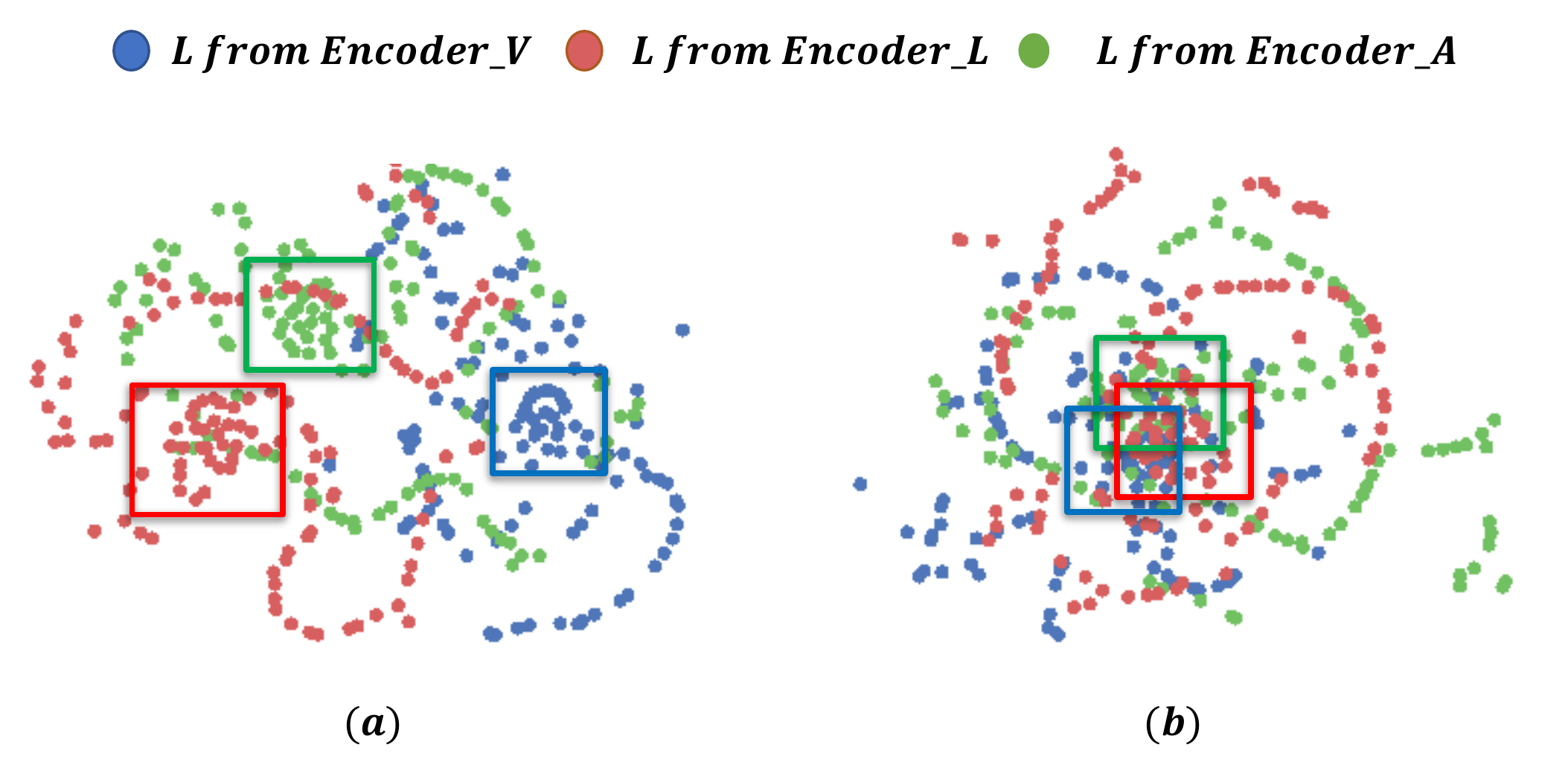}
\caption{T-SNE visualization of encoder output features (a) in the first communication round (b) in the $30^{th}$ communication rounds}
\label{noisy}
\vspace{-0.5em}
\end{figure}

This design has several advantages: 1) The goal of UTMP is predicting from multimodal data, and thus the global model on the server side must have $M$ encoders. We maintain the exact size of model parameters for each client for ease of model uploading and downloading. 2) Intuitively, utilizing outputs from unmatched encoders (e.g., feed $\boldsymbol{X}_L$ to the encoder for $A$) can influence training efficiency and efficacy. However, an intriguing finding is that our model, in fact, utilizes these outputs as noisy signals which prevents client models from overfitting local data. 
However, unlike random noise, these noisy signals carry structural information of sequential input. Although their modalities do not match those of the input data, they can be used as an auxiliary input for preventing overfitting in the following CmDA.
For more evidence, Figure~\ref{noisy} visualizes the latent features with T-SNE from the encoders across the different communication rounds. After multiple runs of training, we observe a shrinkage effect: the extracted features from different encoders become more similar and their distance in latent space shrinks significantly, which represents similar structural information.

\subsection{Hierarchical aggregation}
\subsubsection{Posterior-based Encoder Aggregation (PbEA)} 
The datasets for the same modality are non-IID at local clients, resulting in considerable distributional shifts among local models. 
During the model aggregation, Traditional methods~\cite{mcmahan2017communication, sahu2018convergence} aggregated the models at local clients by averaging the mean of parameters without modeling their uncertainty, which might cause biased prediction.
 Thus, we propose a modality-oriented uncertainty-guided method based on the layer-wise local posteriors to aggregate models trained with data of the same modality using the Markov Chain Monte Carlo (MCMC) method as shown in Algorithm 1. Given the training dataset $(\boldsymbol{X},\boldsymbol{y}) \in D_k$
and a linear model, its least squares loss function  $\ell\left[\upsilon(\boldsymbol{X};\boldsymbol{\theta}_{k}),\boldsymbol{y}\right] = \frac{1}{|D_k|}\|\boldsymbol{X} \boldsymbol{\theta}_k-\boldsymbol{y}\|_{2}^{2}$. Since the square loss corresponds to likelihood under a Gaussian model, the log-likelihood client loss by posteriors becomes:
\begin{equation}
\begin{aligned}
     \ell\left[\upsilon(\boldsymbol{X};\boldsymbol{\theta}_{k}),\boldsymbol{y}\right] & =\log (e^{ \left\{\frac{1}{|D_k|}({\boldsymbol{X} \boldsymbol{\theta} - \boldsymbol{y}})^2  \right\}}) \\ 
    & = \log (e^{  \left\{\frac{1}{|D_k|}\left({\boldsymbol{\theta}_i}-{\mu}_{k}\right)^{\top} {\boldsymbol{\Sigma}}_{k}^{-1}\left({\boldsymbol{\theta}_k}-\boldsymbol{\mu}_{k}\right)\right\}}) + \epsilon
\end{aligned}
\end{equation}
where the mean $\boldsymbol{\mu}_k = (\boldsymbol{X}_k^T\boldsymbol{X}_k)^{-1}\boldsymbol{X}_k^T\boldsymbol{y}_k $, covariance $\boldsymbol{\Sigma}_k^{-1} = \boldsymbol{X}_k^T\boldsymbol{X}_k$, 
and $\epsilon$ denotes a constant.
\begin{algorithm}[]
    \SetAlgoLined
    \SetKwInOut{Input}{input}
    \Input{Sample times $S$, Update steps $T$, Datasets $\{D_0,...,D_K\}$, global model parameter $\boldsymbol{\theta}$}
    \For{$k^{th}$client \textbf{in} K clients}{
    \textbf{Init:} $ SamplesSet = \{\}, \boldsymbol{\theta}_k \xleftarrow{download}  \boldsymbol{\theta}$ \;
        \For{$s \textbf{ in } [1,S]$}{
            Sample $(\boldsymbol{X}_k, \boldsymbol{y}_k) \sim D_k$ \;
            $\boldsymbol{\theta}_k^s = \boldsymbol{\theta}_k$\;
            \For{$t \textbf{ in } [1,T]$}{
                ${\boldsymbol{\theta}_k^s} \gets \boldsymbol{ClientOPT}(\boldsymbol{X}_k, \boldsymbol{y}_k, \boldsymbol{\theta}_k^s)$
                }
                $SamplesSet \cup \{{\boldsymbol{\theta}_k^s}\}$
            }
            \For{$l^{th}$ layer \textbf{in} model}{
            Calculate $\Sigma_k^{l},\mu_k^{l}$ with $SamplesSet$\;
             $\Delta_k^l = {\Sigma}_{k}^{-1}({\boldsymbol{\theta}_{k}^{l}-\mu_{k}^{l}})$
            }
            Record $\boldsymbol{\Delta}_k=[{\Delta}^{1}_k,...,{\Delta}^{L}_k]$ and $\boldsymbol{\Delta}_k \xrightarrow{to} server$
    }
    ${\boldsymbol{\theta}^{\prime}} \longleftarrow \boldsymbol{ServerUpdate}(\boldsymbol{\theta}; \boldsymbol{\Delta}_0, ..., \:\boldsymbol{\Delta}_K)$ \;
    \SetKwInOut{Output}{output}
    \Output{$\boldsymbol{\theta}^{\prime}$}
\caption{Posterior-based Encoder Aggregation}
\end{algorithm}

In federated learning, 
according to the proposition given by \cite{al2020federated}, the global posterior can be calculated by-product of local posteriors as $\mathbb{P}(\boldsymbol{\theta}|(\boldsymbol{X},\boldsymbol{y})) \propto \prod_{k=1}^{K} \mathbb{P}\left({\boldsymbol{\theta}_k}|(\boldsymbol{X}_{k},\boldsymbol{y}_{k})\right)$ where the $K$ represents the number of clients. Accordingly, the mean of global model parameters can be represented by:
\begin{equation}
\begin{aligned}
    \boldsymbol{\mu}:=\left(\frac{1}{K}\sum_{k=1}^{K} \boldsymbol{\Sigma}_{k}^{-1}\right)^{-1}\left(\frac{1}{K}\sum_{k=1}^{K} \boldsymbol{\Sigma}_{k}^{-1} {\boldsymbol{\mu}}_{k}\right)
\end{aligned}
\label{mean}
\end{equation}

Equations (\ref{mean}) requires sending all local means and covariance matrices to the client, which often incurs a high communication and computation burden. To solve this issue, we follow the proposition of global posterior inference~\cite{al2020federated}: $\boldsymbol{\mu}$ is a minimizer of the function $\mathcal{Q}(\boldsymbol{\theta}) := \frac{1}{2}\boldsymbol{\theta}^T (\sum_{k=0}^K \boldsymbol{\sigma}_k^{-1})\boldsymbol{\theta} + (\sum_{k=0}^K \boldsymbol{\sigma}_k^{-1}\boldsymbol{\mu}_k)^T\boldsymbol{\theta}$ whose gradient can be disentangled to local gradients by $\Delta\mathcal{Q}= \sum_{k=1}^{K}\frac{1}{K}\Delta\mathcal{Q}_{k}$ and $ \Delta\mathcal{Q}_{k}=\boldsymbol{\Sigma}_{k}^{-1}({\boldsymbol{\theta}_{k}-\boldsymbol{\mu}_{k}})$.
Accordingly, in order to develop a layer-wise estimation of the global mean $\boldsymbol{\mu}^{l}$, we calculate its gradient by:
\begin{equation}
\begin{aligned}
    \Delta^{l} {:=} K^{-1}{\sum_{k=1}^{K} }({\boldsymbol{\Sigma}}_{k}^{l})^{-1}({\boldsymbol{\theta}_{k}^{l}-\boldsymbol{\mu}_{k}^{l}})
\end{aligned}
\end{equation}

Note that the ${\boldsymbol{\Sigma}}_{k}^{l}$ and $\boldsymbol{\mu}_{k}^{l}$ are the layer-wise covariance and mean. Thus, we obtain $M$ aggregated models consisting of $M$ decoders for CmDA and $M\times M$ encoders for further encoder-decoder concatenation.

\subsubsection{Cross-modal Decoder Aggregation (CmDA)}
 Inspired by Mult\cite{tsai2019multimodal}, which uses the attention mechanism of the Transformer to align the data of different modalities in pairs at the feature level. Since federated learning cannot directly access the data, we consider the alignment at the model parameter level instead. By exploring the correlation between model weights, we implement an attention-alike method\cite{ji2019learning} to align the decoder weights trained on different modality data. Followed by Mult, our proposed cross-modal aggregation strategy takes two decoders' parameters each time from different modalities to compute a self-adaptive coefficient $\psi$, which determines the significance of their hidden signal correlations and can help to facilitate alignment in the modal parameter space. By pairing decoders from all modalities, we get $C_{M}^{2}$ (e.g., for $m\in\{L,A,V\}$,$C_{M}^{2}$=3) optimization objectives. For the $c\in \{1,...,C_{M}^{2}\}$ objective, the corresponding optimization objective can be defined as:

\begin{equation}
\begin{aligned}
    \mathop{arg\thinspace min}\limits_{\boldsymbol{\theta}_{m}^{t}}\frac{1}{2}\psi_{c}^{t}\ast \Gamma(\boldsymbol{\theta}_{m}^{t}, \boldsymbol{\theta}_{\hat{m}}^{t})^{2}
\end{aligned}
\end{equation}
where $m, \hat{m}\in \{L,A,V\} (m\neq\hat{m})$, $\boldsymbol{\theta}_{c}^{t}$ denotes the estimated decoder parameters at communication round $t$, and $\Gamma(,)$ denotes the distance between the model parameters.

We first compute the norm difference between the query $\boldsymbol{\theta}^{t,l}_{m}$ and key $\boldsymbol{\theta}^{t,l}_{\hat{m}}$ in each model layer $l\in \{0,1,...,L\}$ to obtain the layer-wise coefficient $\psi_{c}^{t}=\{\psi_{c}^{t,0}, \psi_{c}^{t,1},..., \psi_{c}^{t,L}\}$:


\begin{equation}
\begin{aligned}
    \psi^{t,l}_{c} = softmax(\gamma^{t,l}_{c}) = softmax(||\boldsymbol{\theta}^{t,l}_{m} - \boldsymbol{\theta}^{t,l}_{\hat{m}}||_{p})
\end{aligned}
\end{equation}



Then, we perform gradient descent to update decoder parameters with the gradients computed by the Euclidean distance for $\Gamma(,)$ and the derivative of Equations (10):

\begin{equation}
\begin{aligned}
    \boldsymbol{\theta}^{t}_{c} \gets \boldsymbol{\theta}^{t}_{m} - \eta\nabla :=\boldsymbol{\theta}^{t}_{m} - \eta\mathop{\sum}\limits_{m=0}^{M-1}\psi_{c}^{t}(\boldsymbol{\theta}^{t}_{m} - \boldsymbol{\theta}^{t}_{\hat{m}})
\end{aligned}
\end{equation}
where $\eta$ is the learning rate. We update the global decoder's parameters by aggregating $\boldsymbol{\theta}^{t}_{c}$ corresponding to the solutions of $C_{M}^{2}$ optimization problems:

\begin{equation}
\begin{aligned}
    \boldsymbol{\theta}^{t+1}_{global} = \frac{1}{C_{M}^{2}}\mathop{\sum}\limits_{c=0}^{C_{M}^{2}-1}\boldsymbol{\theta}^{t}_{c}
\end{aligned}
\end{equation}

In this way, we obtain a global decoder with parameters well represented across the multiple modalities.

\subsection{Encoder-Decoder concatenation}
\begin{figure}[t]
\centering
\includegraphics[scale=0.28]{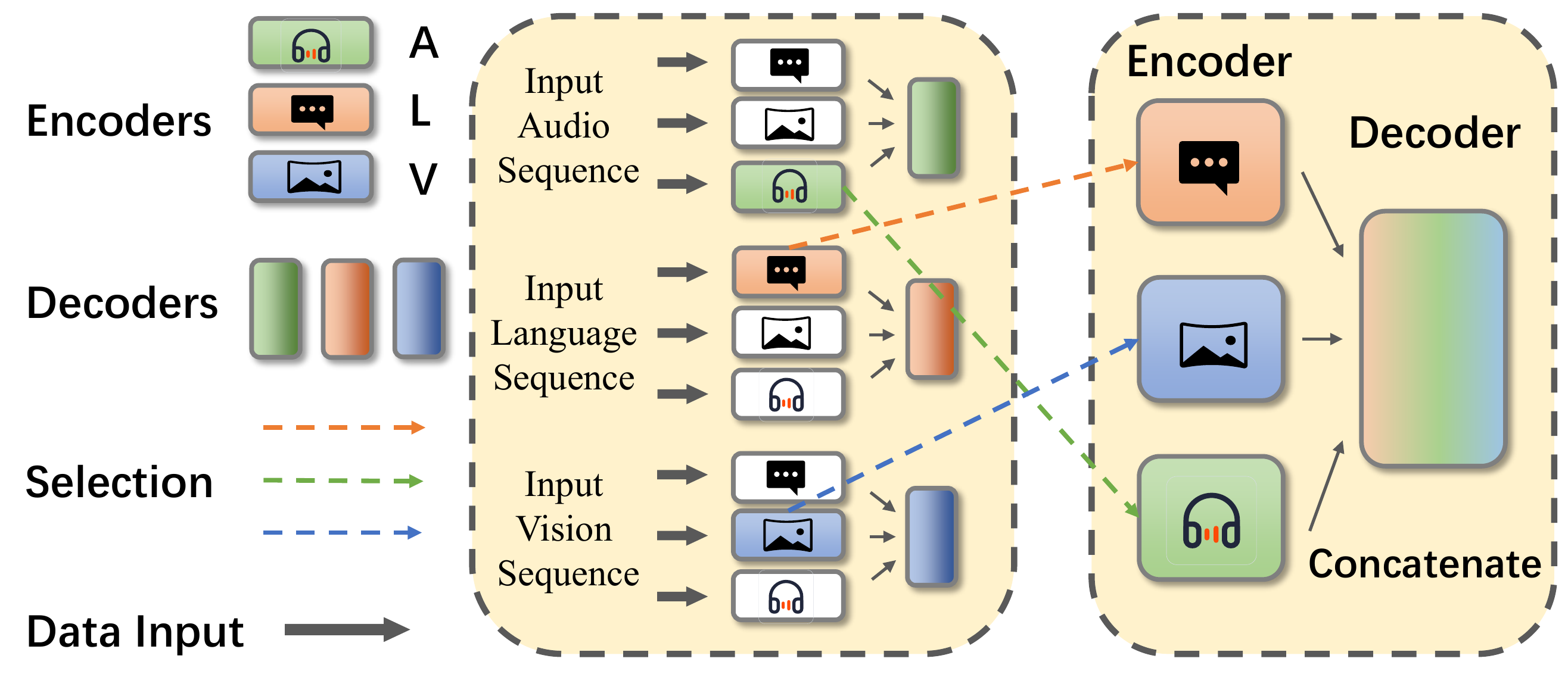}
\caption{Illustration of encoder-decoder concatenation. }
\label{exchange}
\end{figure}

We concatenate the $M\times M$ aggregated encoders and the one global decoder obtained from PbEA and CmDA, which constructs a new model capable of multimodal prediction. To be more specific, as illustrated in Figure \ref{exchange}, after PbEA, each aggregated local model has $M$ ($M=3$ in our example) encoders. Since PbEA is trained with the data in a single modality, we record only the encoder for this modality (i.e., abandoning the rest of $M-1$ encoders) and combine it with the aggregated decoder to be a part of our global multimodal model.

\subsection{Overall Computation Cost}
The training/inference cost of HA-Fedformer is $\mathcal{O}(S^{2}d+k)$/$\mathcal{O}(n)$, while standard FedAvg is $\mathcal{O}(d)$/$\mathcal{O}(n)$, where $S$ indicates the size of sampling set, $k$ represents the number of CmDA, and $d$ is the number of clients.  The model size of Mult is 4.38MB, HA-Fedformer is 846KB, and the communication cost is related to the model size.

\section{Experiments}
\begin{table*}[t]
\centering
  \caption{Results for multimodal sentiment analysis on (relatively
large scale) CMU-MOSEI with unimodal local sequences. $\Downarrow$ means p value of significant test $<$0.01 compare to the HA-Fedformer++. $\pm$ means the variance. $\uparrow$ means higher is better, and $\downarrow$ means lower is better.  Superscript A stands for FedAvg, P for FedProx(e.g., $Mult^{A}$ stands for Mult+FedAvg). HA-Fedformer++ stands for complete HA-Fedformer, HA-Fedformer++(S) is its simplified form, HA-Fedformer+ stands for HA-Fedformer++ minus PbEA, and HA-Fedformer stands for HA-Fedformer+ further minus CmDA.}
\footnotesize
  \label{tab:freq}
  \setlength{\tabcolsep}{2.75mm}{
  \begin{tabular}{|c||ccccc|}
      \toprule

    Metric & $Acc_{7}\uparrow$ & $Acc_{2}\uparrow$ & $F1\uparrow$ & $MAE\downarrow$ & $Corr\uparrow$ \\
    \midrule
    \midrule
        \multicolumn{6}{|c|}{\textbf{CMU-MOSEI Sentiment} (\textcolor{red}{Unimodal local data})}  \\
  \midrule
  \midrule
    $TBJE^{A}$(20' ACL) $\Downarrow$ & 41.5($\pm$3.43E-5) & 68.3($\pm$4.15E-5) & 68.5($\pm$1.14E-5) & 0.843($\pm$5.48E-5) & 0.456($\pm$4.48E-5) \\
    $TBJE^{P}$(20' ACL) $\Downarrow$ & 41.2($\pm$2.25E-5) & 67.4($\pm$1.84E-5) & 68.1($\pm$1.47E-5) & 0.863($\pm$5.15E-5) & 0.477($\pm$7.14E-5) \\
    $MAT^{A}$(20' EMNLP) $\Downarrow$ & 39.4($\pm$2.31E-5) & 63.4($\pm$5.53E-6) & 64.7($\pm$3.14E-6) & 0.857($\pm$8.51E-5) & 0.422($\pm$4.77E-5) \\
    $MAT^{P}$(20' EMNLP) $\Downarrow$ & 39.9($\pm$4.53E-5) & 66.8($\pm$4.22E-5) & 67.3($\pm$2.75E-5) & 0.844($\pm$2.76E-4) & 0.456($\pm$6.69E-5) \\
    $MNT^{A}$(20' EMNLP) $\Downarrow$ & 38.8($\pm$1.70E-4) & 63.2($\pm$1.53E-4) & 64.6($\pm$8.93E-5) & 0.871($\pm$1.47E-5) & 0.376($\pm$1.31E-4) \\
    $MNT^{P}$(20' EMNLP) $\Downarrow$ & 36.0($\pm$5.54E-5) & 63.1($\pm$3.82E-5) & 64.7($\pm$2.23E-5) & 0.964($\pm$7.47E-6) & 0.435($\pm$6.44E-4) \\
    $Mult^{A}$(19' ACL) $\Downarrow$ & 42.7($\pm$1.49E-5) & 69.0($\pm$2.13E-5) & 72.7($\pm$2.26E-4) & 0.783($\pm$1.88E-5) & 0.374($\pm$1.30E-3) \\
    $Mult^{P}$(19' ACL) $\Downarrow$ & 41.9($\pm$6.52E-6) & 65.5($\pm$7.43E-6) & 70.4($\pm$8.94E-5) & 0.806($\pm$4.96E-6) & 0.269($\pm$8.33E-5) \\
    $FedMSplit$(22' KDD) $\Downarrow$ & 43.8($\pm$7.53E-5) & 73.5($\pm$8.42E-5) & 75.2($\pm$7.79E-5) & 0.702($\pm$6.96E-5) & 0.522($\pm$9.41E-5) \\
      \midrule
    \midrule
      \multicolumn{6}{|c|}{\textbf{CMU-MOSEI Sentiment} (\textcolor{blue}{Ablation Study})}  \\
          \midrule
    \midrule
    only L \& A(ours) $\Downarrow$ & 47.8($\pm$4.22E-5) & 78.4($\pm$7.45E-5) & 78.1($\pm$8.93E-6) & 0.644($\pm$2.37E-5) & 0.614($\pm$1.75E-5) \\
    only V \& L(ours) $\Downarrow$ & 46.5($\pm$8.95E-6) & 78.0($\pm$6.34E-5) & 78.0($\pm$2.26E-5) & 0.656($\pm$1.15E-5) & 0.606($\pm$9.83E-6) \\
    only A \& V(ours) $\Downarrow$ & 42.3($\pm$1.23E-6) & 63.0($\pm$4.37E-5) & 76.4($\pm$3.24E-5) & 0.803($\pm$2.28E-5) & 0.195($\pm$1.42E-5) \\
  \midrule
    \midrule
    HA-Fedformer $\Downarrow$ & 45.4($\pm$3.33E-5) & 77.7($\pm$3.71E-5) & 78.7($\pm$5.12E-7) & 0.662($\pm$3.26E-5) & 0.604($\pm$3.83E-5) \\
    HA-Fedformer+ $\Downarrow$ & 46.7($\pm$3.72E-5) & 78.0($\pm$2.84E-5) & 78.4($\pm$9.76E-6) & 0.647($\pm$1.91E-5) & \textbf{0.625($\pm$4.14E-5)} \\
    HA-Fedformer++(S) $\Downarrow$ & \textbf{49.1($\pm$4.42E-5)} & 78.5($\pm$2.74E-5) & 78.9($\pm$3.72E-5) & 0.639($\pm$1.26E-5) & 0.617($\pm$2.46E-5) \\
    HA-Fedformer++ & 48.6($\pm$1.92E-6) & \textbf{79.1($\pm$2.25E-5)} & \textbf{79.2($\pm$2.92E-5)} & \textbf{0.638($\pm$7.43E-6)} & 0.624($\pm$3.94E-6) \\
  \bottomrule
\end{tabular}}
\label{mosei}
\end{table*}

In this section, we empirically evaluate \textit{HA-Fedformer} on two well-known datasets that are frequently used to benchmark the multimodal sentiment analysis in prior works.~\cite{wang2019words, pham2019found, tsai2019multimodal}. Our goal is to compare \textit{HA-Fedformer} with previous SOTA baselines under the context of data non-IID and \textit{unaligned} multimodal language sequences under the UTMP framework. 

\subsection{Datasets and evaluation metrics}
\textbf{CMU-MOSI and CMU-MOSEI} 
CMU-MOSI~\cite{zadeh2016multimodal} is a multimodal sentiment analysis dataset consisting of 2,199 short monologue video clips, while CMU-MOSEI~\cite{zadeh2018multimodal} consists of 23,454 movie review video clips from YouTube. Each video clip's length is equivalent to the length of a sentence.
We evaluate the performance with the following metrics (in accordance with those employed in previous works~\cite{wang2019words, tsai2019multimodal, pham2019found}): 7-class accuracy (i.e., $Acc_{7}\in[-3,3]$), binary accuracy (i.e., $Acc_{2}\in\{negative, positive\}$), F1 score, mean absolute error (MAE),  and the correlation of the model’s prediction with human (Corr). It should be noted that the dataset is distributed equally among all clients, while 10 out of 30 clients participate in FL in each communication round.



\begin{figure*}[t]
\centering
\includegraphics[scale=0.7]{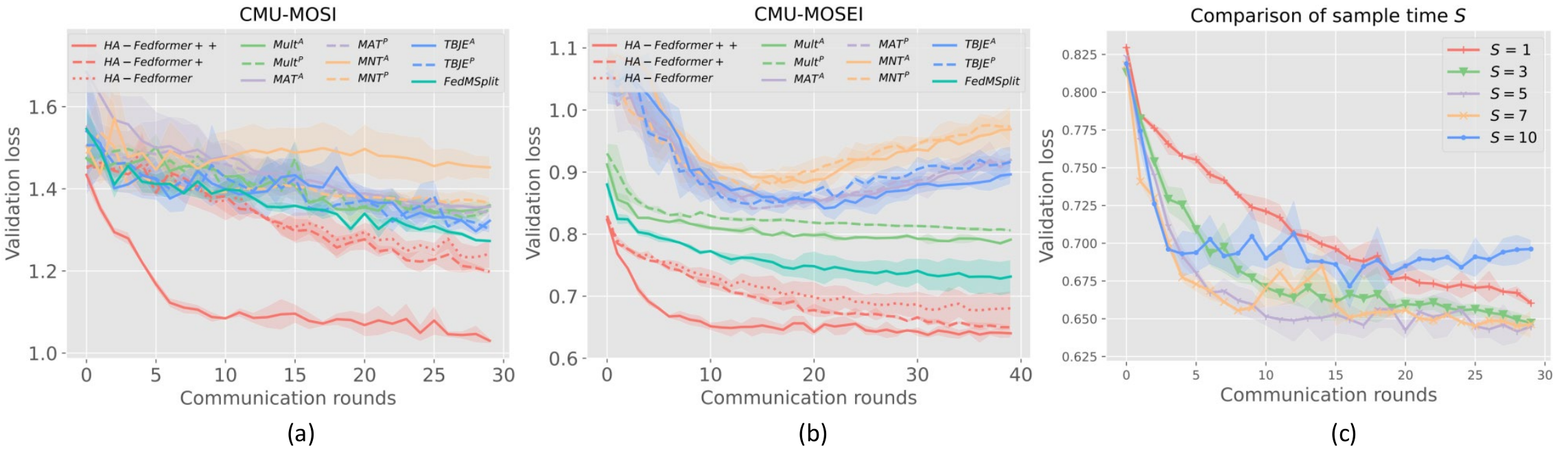}
\caption{(a),(b) Training loss $v.s.$ Communication rounds for two datasets. (c) Comparison of sample times $S$. It should be noted that the same baseline uses the same color, and different dashed lines represent different federated learning methods. All representations follow the same as Table 1.}
\vspace{-1em}
\label{eval}
\end{figure*}

\subsection{Baselines}

We choose Multimodal Transformer (Mult)~\cite{tsai2019multimodal}, Transformer-based Joint Encoding (TBJE)~\cite{delbrouck2020transformer}, Modulated Normalization Transformer (MNT), Modulated Attention Transformer (MAT)~\cite{delbrouck2020transformer}, and FedMSplit~\cite{chen2022fedmsplit} that achieves SOTA results on various multimodal learning tasks as our baselines. We further combine two SOTA federated learning approaches, FedAvg~\cite{mcmahan2017communication} and FedProx~\cite{sahu2018convergence} with each non-federated baseline in order to extend them to the federated learning context. We use superscript $A$ to denote FedAvg and $P$ to denote FedProx (e.g., $Mult^{A}$ represents Mult+FedAvg). HA-Fedformer++ stands for complete HA-Fedformer, HA-Fedformer+ stands for HA-Fedformer++ minus PbEA, and HA-Fedformer stands for HA-Fedformer+ further minus CmDA. We further provide a simplified solution HA-Fedformer++(S) (refer to Appendix A.4.) for HA-Fedformer++, that only takes advantage of the mean value of the sampling results.

\begin{table}[t]
\centering
  \caption{Results for multimodal sentiment analysis on  CMU-MOSI with unimodal data. All representations follow the same as in Table 1. Refer to Appendix A.2. for further results.}
\footnotesize
  \label{tab:freq}
  \resizebox{0.475\textwidth}{!}{
  \setlength{\tabcolsep}{0.9mm}{
  \begin{tabular}{|c||ccc|}
      \toprule

    Metric & $Acc_{7}\uparrow$ & $Acc_{2}\uparrow$  & $Corr\uparrow$ \\
    \midrule
    \midrule
        \multicolumn{4}{|c|}{\textbf{CMU-MOSI Sentiment} (\textcolor{red}{Unimodal local data})}  \\
  \midrule
  \midrule
    $TBJE^{A}$ $\Downarrow$ & 26.1($\pm$3.3E-5) & 69.4($\pm$3.2E-5)  & 0.38($\pm$7.1E-5)  \\
    $TBJE^{P}$ $\Downarrow$ & 26.6($\pm$1.1E-5) & 69.7($\pm$4.2E-5)  & 0.37($\pm$6.5E-5) \\
    $MAT^{A}$ $\Downarrow$ & 25.2($\pm$5.4E-5) & 68.5($\pm$4.8E-5)  & 0.37($\pm$7.1E-5) \\
    $MAT^{P}$ $\Downarrow$ & 25.6($\pm$2.4E-4) & 68.0($\pm$3.2E-5)  & 0.34($\pm$2.7E-5) \\
    $MNT^{A}$ $\Downarrow$ & 22.0($\pm$4.7E-5) & 67.8($\pm$3.7E-5)  & 0.38($\pm$1.5E-5) \\
    $MNT^{P}$ $\Downarrow$ & 22.8($\pm$3.6E-5) & 67.4($\pm$2.7E-5)  & 0.36($\pm$8.3E-5) \\
    $Mult^{A}$ $\Downarrow$ & 23.5($\pm$6.3E-5) & 63.7($\pm$3.2E-4)  & 0.37($\pm$5.3E-4) \\
    $Mult^{P}$ $\Downarrow$ & 22.2($\pm$1.2E-4) & 64.6($\pm$3.3E-5)  & 0.39($\pm$5.5E-4) \\
    $FedMSplit$ $\Downarrow$ & 27.7($\pm$2.3E-4) & 68.9($\pm$3.3E-4)  & 0.44($\pm$4.77E-4) \\
      \midrule
    \midrule
      \multicolumn{4}{|c|}{\textbf{CMU-MOSI Sentiment} (\textcolor{blue}{Ablation Study})}  \\
          \midrule
    \midrule
    L \& A $\Downarrow$ & 28.1($\pm$1.7E-6) & 72.5($\pm$5.3E-4)  & 0.55($\pm$9.6E-4) \\
    V \& L $\Downarrow$ & 29.3($\pm$6.3E-5) & 73.5($\pm$3.1E-4)  & 0.55($\pm$1.3E-3) \\
    A \& V $\Downarrow$ & 17.8($\pm$3.7E-5) & 52.7($\pm$4.4E-5)  & 0.17($\pm$3.3E-4) \\
  \midrule
    \midrule
    HA-Fed. $\Downarrow$ & 29.0($\pm$1.0E-4) & 72.0($\pm$1.6E-5)  & 0.51($\pm$9.8E-5) \\
    HA-Fed.+ $\Downarrow$ & 29.3($\pm$3.5E-5) & 72.1($\pm$4.3E-5)  & 0.52($\pm$5.8E-4) \\
    HA-Fed.++(S)$\Downarrow$ & 30.7($\pm$9.2E-5) & 74.8($\pm$1.5E-5)  & 0.54($\pm$1.2E-5) \\
    HA-Fed.++ & \textbf{31.1($\pm$2.3E-6)} & \textbf{76.3($\pm$6.2E-5)}  & \textbf{0.55($\pm$7.7E-5)} \\
  \bottomrule
\end{tabular}}
}
\label{mosi}
\vspace{-0.5cm}
\end{table}

\subsection{Benchmark results}
We evaluate \textit{HA-Fedformer} on two datasets, and the results are shown in Table \ref{mosei}, \ref{mosi}. 
Traditional multimodal learning methods applying FL can hardly achieve satisfactory results under the UTMP, e.g., the averaged $Corr$ for CMU-MOSEI and CMU-MOSI are $0.408$ and $0.376$, respectively.
In comparison, \textit{HA-Fedformer} can get a satisfying $Corr$ (e.g., $Corr$=0.625) on every benchmark dataset. The improvement upon the baseline methods is between 15$\%$-20$\%$ under a majority of evaluation metrics. Meanwhile, \textit{HA-Fedformer} can even achieve the result of $acc_{7}$\textgreater50.

Figure \ref{eval} (a) and (b) illustrate the validation losses of baseline methods under two datasets. We observe that it is difficult for traditional multimodal learning methods to converge under UTMP. Especially when using CMU-MOSEI, some baselines exhibit serious overfitting issues. While our proposed \textit{HA-Fedformer} can stably reduce the validation loss until the model converges.


We also measure the statistical significance of the results with
one-tailed Wilcoxon’s signed-rank test~\cite{wilcoxon1992individual}, and the test result is shown in Table 1.
Each method is compared with HA-Fedformer++, and $\Downarrow$ denotes 'significantly worse' with p$<$0.01. The test shows that our approach is significantly better than others, both on source and target.

\subsection{Ablation study}

We further conduct a comprehensive ablation analysis with two benchmark datasets. The results are shown in Table \ref{mosei}, \ref{mosi}, and Figure \ref{eval} (c).

First, we study the influence of data modalities on the performance baseline methods. We take L and A, V and L, A and V as local input data to train \textit{HA-Fedformer}. We can observe from the bottom of Table \ref{mosei}, \ref{mosi} that reducing the data of one modality will degrade the performance of the model, and the impact of removing the (L) data is the most obvious. Even using only two modalities of data, our method outperforms the vast majority of baselines, which further demonstrates the superiority of our method.

Second, we study the importance of hierarchical aggregation. From Table \ref{mosei}, \ref{mosi}, we can observe that after applying the hierarchical aggregation, all metrics can be significantly improved, especially for $Acc_{7}$ and $Corr$. This shows that the hierarchical aggregation can substantially alleviate the data non-IID and unaligned data sequences issues by providing a better representation of multi-modal data.

Finally, we examine the influence of choosing different sample times $S$ during PbEA, and the result is shown in Figure \ref{eval} (c). Following \cite{lakshminarayanan2017simple}, the default value of $S$ is set to 5, and we explore using different $S$ for training the model. We can observe that as $S$ increases, the model converges faster. However, when $S=10$, the training process falls into overfitting. We find the model obtains its best performance when $S=5$ and $S=7$. Considering the resource consumption, we empirically set $S=5$ throughout the experiments.

\begin{table}[t]
\centering
  \caption{Results for multimodal sentiment analysis on CMU-MOSEI under different modality Missing Rate (MR). All representations follow the same as in Table 1.}
\footnotesize
  \label{tab:freq}
  \resizebox{0.475\textwidth}{!}{
  \setlength{\tabcolsep}{0.9mm}{
  \begin{tabular}{|c||ccc|}
      \toprule

    MR ($Acc_{7}$) & $\Xi=0.7$ & $\Xi=0.5$  & $\Xi=0.3$ \\
    \midrule
    \midrule
        \multicolumn{4}{|c|}{\textbf{CMU-MOSEI Sentiment} (\textcolor{blue}{Ablation Study})}  \\
  \midrule
  \midrule
    FedMSplit& 41.6($\pm$1.7E-5) & 41.8($\pm$2.1E-5)  & 42.5($\pm$4.3E-5)  \\
    HA-Fed.++& 43.6($\pm$2.1E-4) & 44.3($\pm$2.4E-4)  & 47.8($\pm$7.8E-5)  \\
  \bottomrule
\end{tabular}}
}
\label{robust}
\end{table}

\subsection{Robustness analysis}
We also examine the robustness of HA-Fedformer with CMU-MOSEI when some modalities are missing.
We let $\Xi_{J}$ to indicate Missing Rate (MR) suggesting the probability a client does not have the modality-$j$ for inference. We set equal missing rates for each modality $\Xi_{1}=\Xi_{M}=\Xi$. As shown in Tab. \ref{robust}, HA-Fedformer still outperforms SOTA method FedMSplit~\cite{chen2022fedmsplit} under every scenario and is even competitive with FedMSplit when $\Xi=0$ ($Acc_{7}$=43.8). 


\section{Conclusion and limitations}
In this paper, we proposed a Multimodal Federated Transformer with Hierarchical Aggregation. Unlike prior approaches that used multimodal data as local input, \textit{HA-Fedformer} for the first time solved the multimodal FL problem under the \textit{Unimodal Training - Multimodal Prediction} framework. Together with PbEA and CmDA, we further boost the performance of \textit{HA-Fedformer} to the point where it is compatible with SOTA multimodal works (i.e., not an FL setting). Although the uncertainty estimation in PbEA brings considerable computational cost and extra time consumption,
we believe \textit{HA-Fedformer}, with only 846KB,  opens up a new path for multimodal federated learning, making it no longer limited by the size or data modality. 



{\small
\bibliographystyle{ieee_fullname}
\bibliography{egbib}
}

\end{document}